%% file: emnlp2021.tex
\pdfoutput=1

\documentclass[11pt]{article}

\usepackage{emnlp2021}

\usepackage{times}
\usepackage{latexsym}

\usepackage[T1]{fontenc}
\usepackage[utf8]{inputenc}

\usepackage{microtype}

\usepackage{graphicx}
\usepackage{booktabs}
\usepackage{enumitem}

\usepackage{algorithm}
\usepackage{algpseudocode}

\usepackage{url}

\definecolor{mycolor}{RGB}{229, 83, 0}

\newcommand{\DATASETNAME}{NewsCLIPpings Dataset}
\newcommand{\DATASETABBR}{NewsCLIPpings}

\newcommand{\myparagraph}[1]{\vspace{0.2cm}\noindent\textbf{\emph{#1}}}
\newcommand{\boldparagraph}[1]{\vspace{0.2cm}\noindent\textbf{#1}}

\title{NewsCLIPpings: Automatic Generation of \\Out-of-Context Multimodal Media}

\author{Grace Luo
\And
Trevor Darrell \\ \\
{University of California, Berkeley} \\
{\tt\small \{graceluo, trevordarrell, anna.rohrbach\}@berkeley.edu}
\And
Anna Rohrbach \\
}

\begin{document}
\maketitle

\begin{abstract}
Online misinformation is a prevalent societal issue, with adversaries relying on tools ranging from cheap fakes to sophisticated deep fakes. We are motivated by the threat scenario where an image is used out of context to support a certain narrative. While some prior datasets for detecting image-text inconsistency generate samples via text manipulation, we propose a dataset where both image and text are unmanipulated but \emph{mismatched}. We introduce several strategies for automatically retrieving convincing images for a given caption, capturing cases with inconsistent entities or semantic context. Our large-scale automatically generated \DATASETNAME{}: (1) demonstrates that machine-driven image repurposing is now a realistic threat, and (2) provides samples that represent challenging instances of mismatch between text and image in news that are able to mislead humans. We benchmark several state-of-the-art multimodal models on our dataset and analyze their performance across different pretraining domains and visual backbones.
\end{abstract}

\input{sections/intro}

\input{sections/related}

\input{sections/dataset}
\input{sections/experiments}
\input{sections/additional}
\input{sections/conclusion}
\myparagraph{Acknowledgements.}
This work was supported in part by DoD including DARPA's XAI, LwLL, and/or SemaFor programs, as well as BAIR's industrial alliance programs.

\clearpage
\input{sections/ethics}
\bibliography{anthology,custom}
\bibliographystyle{acl_natbib}
\clearpage
\input{sections/appendix}

\end{document}

%% file: sections/intro.tex
\section{\label{sec:intro}Introduction}

Misinformation has reached new heights as sophisticated AI-based tools have come into the spotlight. For instance, it has become easy to generate images of people who ``do not exist''\footnote{https://thispersondoesnotexist.com} %
and create realistic deepfakes of existing people ~\cite{deepfakes}.
Recent language models have become better at fooling people into believing that generated texts are from real people ~\cite{gpt3blog}.
However, simple and cheap image repurposing remains one of the most widespread and effective forms of misinformation~\cite{outofcontext}. %
Specifically, real images of people and events get reappropriated and used out of context to illustrate false events and misleading narratives by misrepresenting \emph{who} is in the image, what is the \emph{context} in which they appear, or \emph{where} the event takes place. This method is effective since augmenting a story with an image has been shown to increase user engagement and make false stories seem true~\cite{fenn2019nonprobative}.
Here, we explore whether such a threat can be \emph{automated}. 
We show that real world images can be automatically matched to captions to generate false but compelling news stories, a threat scenario that may lead to larger-scale image repurposing.  

\begin{figure}[t]
\begin{center}
\includegraphics[width=\linewidth]{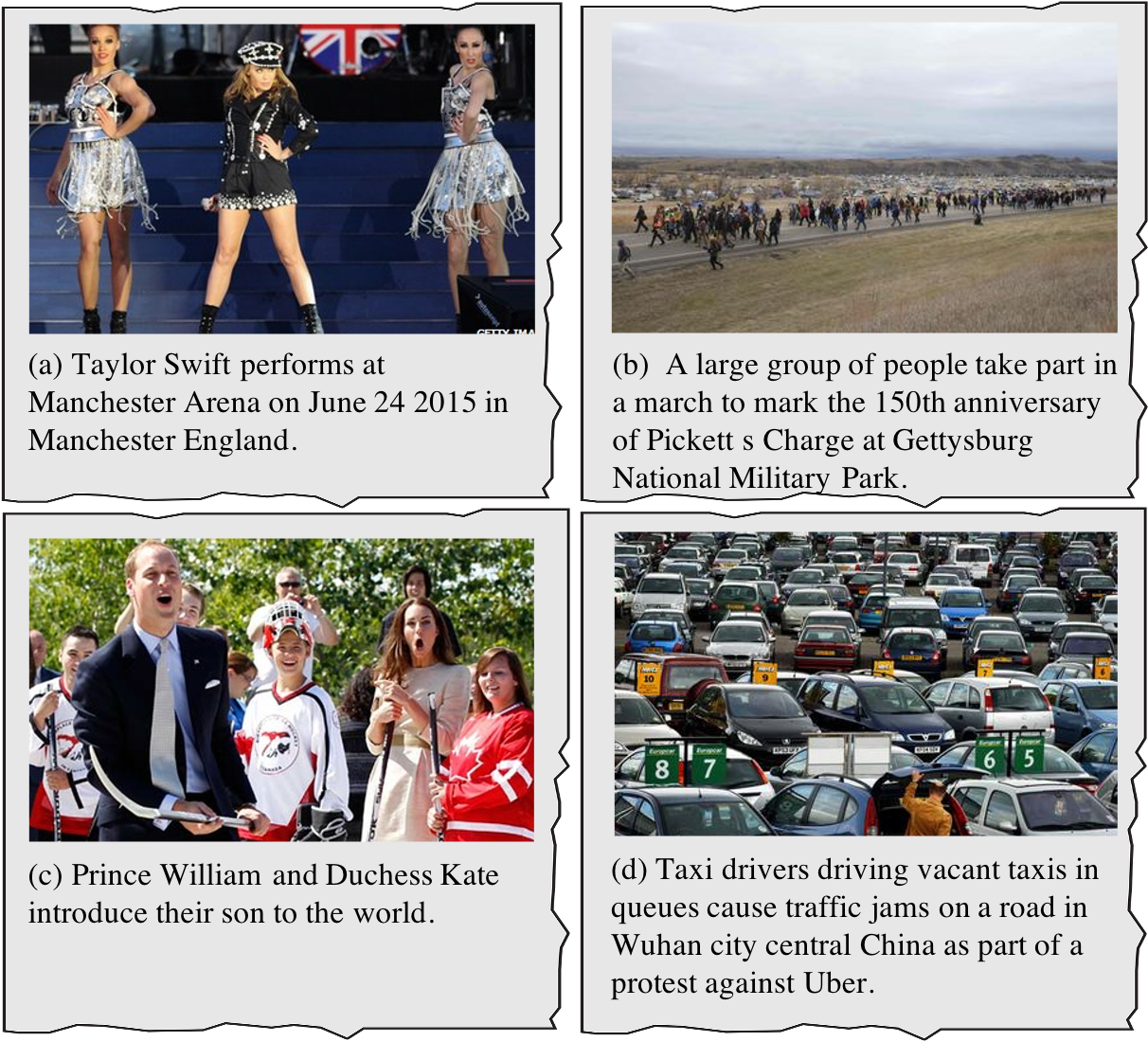} %
\vspace{-0.5cm}
\caption{Consider the following examples and guess whether these are pristine news or automatically matched image-caption pairs. The solution and more discussion are given in text.}
\label{fig:teaser}
\vspace{-0.8cm}
\end{center}
\end{figure}

While synthetic media conceptually could be detected by doing unimodal analysis (e.g. a detector for GAN-generated images), in our case both the text and the image are real. Thus, determining whether an image-caption pair is pristine or falsified requires joint multimodal analysis of the image and text (consider Figure~\ref{fig:teaser} and make your guess).

Prior work has proposed several datasets related to our problem statement. One line of work obtains out-of-context image-text pairs by manipulating the named entities within the text ~\cite{muller2020multimodal,sabir2018deep}. We find that in practice this may lead to linguistic inconsistencies, providing sufficient signal for a text-only model to distinguish between pristine and falsified descriptions without looking at the images.
One recent work on detecting out-of-context images~\cite{aneja2021catching} focuses on a scenario where an image is accompanied by two captions (from two distinct news sources), and one has to establish whether the two captions are consistent. Here, we do not manipulate textual descriptions as we aim to minimize unimodal bias in our task. We do not assume that two captions are available per image, rather we focus on classifying each image-caption pair as pristine or falsified.

Specifically, we propose a large-scale automatically constructed dataset with real and out-of-context news based on the VisualNews~\cite{liu2020visualnews} corpus. We consider several threat scenarios, designing matches based on: (a) \emph{caption-image similarity}, (b) \emph{caption-caption similarity}, where we retrieve an image with similar semantics to a given caption while the named entities between the source and the target are disjoint,  
(c) \emph{person match}, where  we retrieve an image that depicts a person mentioned in the source caption but pictured in a different context, %
and (d) \emph{scene match}, where we retrieve an image that has the same scene type as the source image but depicts a different event
\footnote{This answers the question in Figure~\ref{fig:teaser}, i.e., examples a), b), c), d) correspond to these four threat scenarios in our dataset, so all four are falsified.}. 
We use the recent powerful multimodal model CLIP~\cite{radford2learning} and other image and text models to construct the \textbf{ \DATASETNAME{}}. To make our dataset more challenging, we introduce an adversarial filtering technique based on CLIP.

We benchmark several state-of-the-art multimodal models and analyze their performance on the \DATASETNAME{}. We investigate the impact of the pretraining domains and various visual backbones. We conduct a human evaluation that shows humans find it challenging to distinguish between pristine and falsified samples from our dataset. We also perform a qualitative analysis with the help of visual salience to shed light onto the useful cues discovered by the models trained on our dataset. Our dataset is publicly available here: \href{https://github.com/g-luo/news\_clippings}{https://github.com/g-luo/news\_clippings} \footnote{Specifically, we provide pristine and falsified matches for captions/images, i.e. their identifiers within the VisualNews dataset. The copyright and usage rights of the data are subject to that of~\cite{liu2020visualnews}.}.

%% file: sections/related.tex
\section{\label{sec:related}Related Work}
We review several most relevant datasets in detail. %

Some earlier proposed datasets for detecting multimodal misinformation are MultimodAl Information Manipulation dataset (MAIM)~\cite{jaiswal2017multimedia} and Multimodal Entity Image Repurposing (MEIR)~\cite{sabir2018deep}. MAIM naively matches images to captions from other random images to create their falsified versions. MEIR introduces swaps over named entities for people, organizations and locations. One of their assumptions is that for each image-caption ``package'' there is an unmanipulated related package (geographically near and semantically similar) in the reference set. This allows verifying the integrity of the query package by first retrieving a related package and then comparing the two. This problem statement is different from ours, as we do not assume availability of a perfect reference set.
A more recent work has proposed TamperedNews~\cite{muller2020multimodal}, a dataset where named entities specific to people, locations and events are swapped to other random\footnote{With some constraints, such as individuals of the same country and gender or locations within the same region.} named entities within the article body. We show that such text manipulations lead to significant linguistic biases and the corresponding tasks can be solved without looking at the images (See Section ~\ref{sec:experiments} for more details).

Another recent work~\cite{aneja2021catching} aims to detect when images are used out of context, somewhat similar to MEIR above. They collect a dataset where each image appears in two distinct news sources and thus is associated with two captions. Most of the collected data is not labeled, but a small subset has been manually annotated as in- or out-of-context. Their problem statement (analyzing image and two captions) is again different from ours.

One other work~\cite{tan2020detecting} tackles Neural News generation by replacing real articles with Grover~\cite{zellers2019defending} generated text and real captions with synthetic ones. They do not mismatch the images, which remain relevant to the article's content. The impact of image analysis on this task is rather limited, while analyzing the captions and the article body is key to the best detection performance.

Finally, some work focuses on human-made fake news detection, such as FakenewsNet~\cite{shu2020fakenewsnet} and Fakeddit~\cite{nakamura2019r}, etc. While these datasets contain important real world examples of fake news, our focus is on exploring an \emph{automated} threat scenario, where an image is automatically retrieved to match a given caption.

%% file: sections/dataset.tex
\section{\label{sec:dataset}The \DATASETNAME{}}

\begin{figure}[t]
\begin{center}
\includegraphics[width=\linewidth]{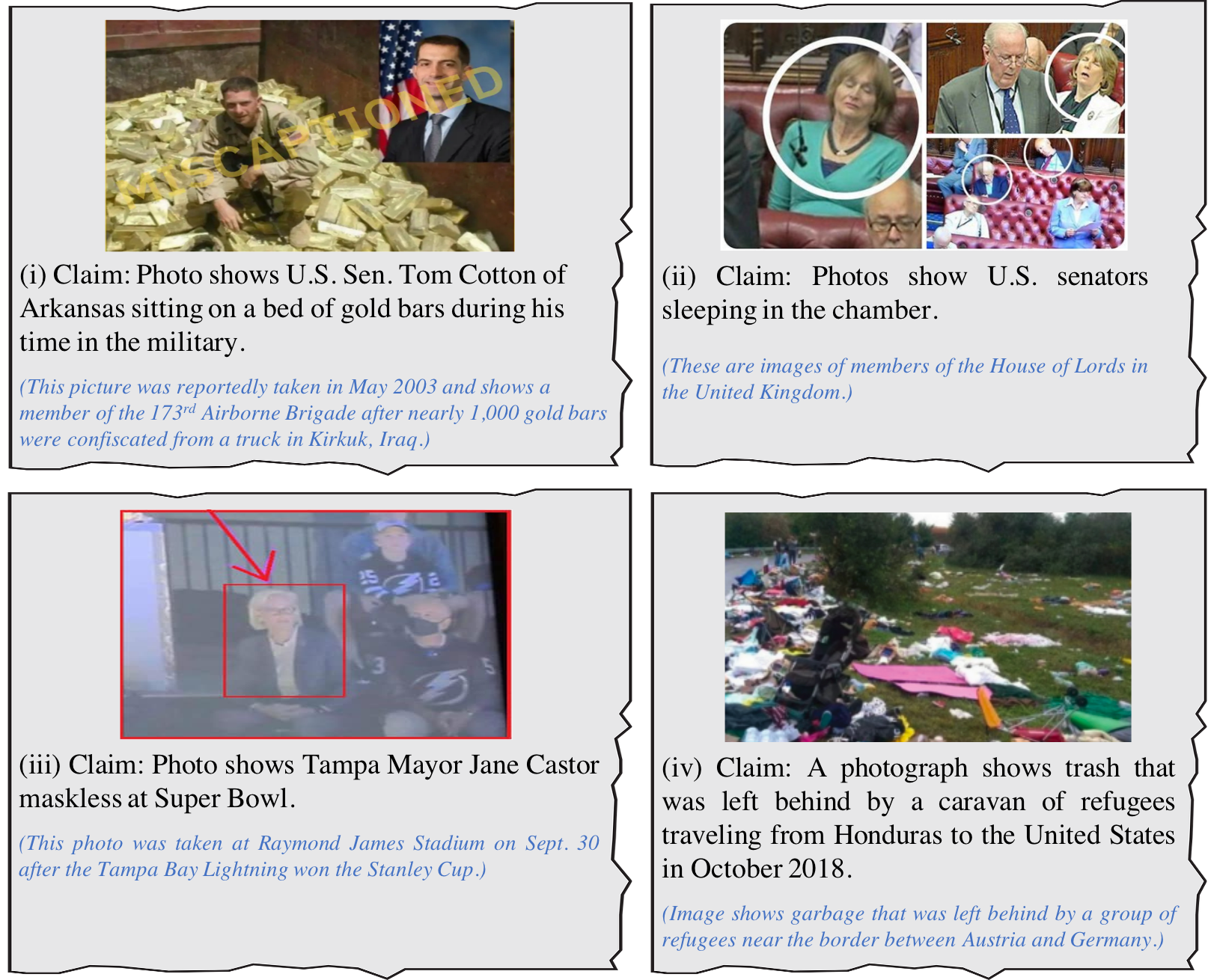}
\vspace{-0.5cm}
\caption{We are motivated by the real-world examples of images used out-of-context. Here we include \emph{real} misinformation examples found online\textsuperscript{\ref{ftn:real}} which closely resemble the four threat scenarios in our dataset.}
\label{fig:real-world}
\vspace{-0.8cm}
\end{center}
\end{figure}

The objective of this work is to explore techniques for creating challenging, non-random image-caption matches that require fine-grained semantic and entity knowledge. As seen in Figure~\ref{fig:real-world}\footnote{\label{ftn:real}Examples found on \url{https://www.snopes.com} and \url{https://www.politifact.com}.}, misinformation in the wild is often extremely subtle and much more difficult than the random matches provided in prior synthetic datasets. In fact, general models that were not specifically trained or finetuned on the news domain can ``solve'' random news matches. We found that CLIP was able to achieve $97.39\%$ Top-1 accuracy on a caption-image retrieval task with news images\footnote{We ran this on a random 40k subset of VisualNews and counted how often CLIP selected the true image vs. four random negative images.}. For comparison, a recent method TRIP~\cite{thomas2020preserving} reports a Top-1 accuracy of $73.78\%$ on a similar task. As a result, we construct several splits that model specific threat scenarios seen in the real world, and we use CLIP ViT-B/32 \textit{off-the-shelf} to filter out the less challenging samples.

In the following, we assume we have a pristine query pair $(img_1,cap_1)$ and retrieve another pair $(img_2, cap_2)$ to form a falsified pair $(img_2,cap_1)$.

\myparagraph{Preprocessing} Our dataset is derived from VisualNews~\cite{liu2020visualnews}, a large-scale corpus which contains image-caption pairs from four news agencies (The Guardian, BBC, USA Today, and The Washington Post). We use spaCy NER~\cite{honnibal2017spacy} to label named entities in captions and the Radboud Entity Linker (REL)~\cite{vanHulst:2020:REL} to link them to their Wikipedia 2019 entries.
We compute text embeddings using SBERT-WK~\cite{wang2020sbert} and CLIP~\cite{radford2learning}. We compute image embeddings with Faster R-CNN~\cite{ren2015faster} and CLIP. We use a ResNet50 classifier trained on the Places365 dataset~\cite{zhou2017places} to get scene embeddings from images. We ensure that all matched samples are at least 30 days apart and that $(cap_1,cap_2)$ have no overlapping named entities identified by spaCy and REL to prevent true matches, with the exception of the Person split, where we expect at least one ``PERSON'' entity to match.

\myparagraph{Query for Semantics} Our first split models a threat scenario that queries for specific semantic content, with the intent to portray the subjects of the image as certain other named entities, see Figure~\ref{fig:real-world} (i, ii). We consider two ways of getting the matches. \textbf{(a) CLIP Text-Image}: We rely on the state-of-the-art CLIP representation to retrieve samples with the highest CLIP text-image similarity between $(img_2,cap_1)$. \textbf{(b) CLIP Text-Text}: We match samples with the highest CLIP text-text similarity between $(cap_1,cap_2)$ and retrieve the corresponding $img_2$. See examples (a) and (b) in Figure~\ref{fig:teaser}.

\myparagraph{Query for Person} This split models a threat scenario that queries for a specific person, with the intent to portray them in a false context, as in Figure~\ref{fig:real-world} (iii). We ensure that the person of interest is pictured: all considered samples must have ``PERSON'' entities in their captions and a person related Faster-RCNN bounding box detected in the image. To avoid cases where the query person is mentioned but unlikely to be pictured, we filter captions where the person is in the possessive form, the object of the sentence, or modify a noun as determined by spaCy's dependency parser.
We ensure that the context is distinct:
the Places365 ResNet similarity must be less than $0.9$. Finally, we found that there were a number of unsolvable falsified samples where the caption could be plausibly matched with any image of the person of interest. We minimize the number of such ``generic'' captions: we finetune a BERT~\cite{devlin2019bert} model on a small labelled subset of our training data to filter these captions from our matching process.
\textbf{(c) SBERT-WK Text-Text}: We match samples that mention the query person based on the lowest semantic similarity measured by their SBERT-WK score, a text-only sentence embedding. See example (c) in Figure~\ref{fig:teaser}.

\begin{figure*}[t]
\begin{center}
\includegraphics[width=\linewidth]{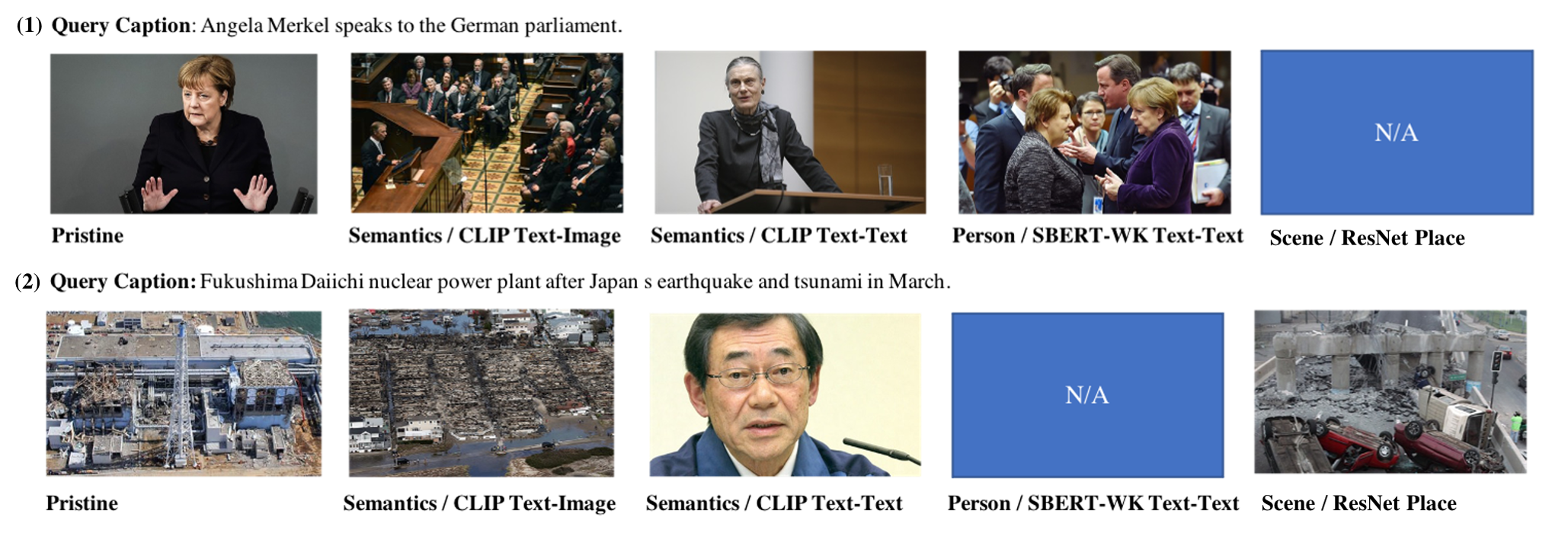}
\vspace{-0.6cm}
\caption{Comparison of the retrieved matches for the same query caption obtained within our four splits. %
}
\label{fig:supplemental-comparison}
\vspace{-0.5cm}
\end{center}
\end{figure*}

\myparagraph{Query for Scene} This split models a threat scenario that queries for a specific scene, with the intent to mislabel the event, see Figure~\ref{fig:real-world} (iv). All samples must have no ``PERSON'' named entities in the captions. This aims to filter headshots and other images with little scene information. 
\noindent\textbf{(d) ResNet Place: } We match samples with the highest Places365 image similarity, as determined by the dot product of their ResNet embeddings.
See example (d) in Figure~\ref{fig:teaser}.

\myparagraph{Merged Split} This split mixes samples from all the splits to model a more realistic case where a  variety of methods are used to generate out-of-context images, i.e. all types of mismatch may be encountered at test time. We merge the splits such that there is an equal number of samples from every split, and the captions and images across splits are disjoint. 

\myparagraph{Adversarial CLIP Filtering} 
In the initial version of our dataset, we observed a distributional shift of CLIP Text-Image scores between the pristine $(img_1,cap_1)$ and falsified $(img_2,cap_1)$ samples. This makes sense, since it is not always possible to find a falsified image that is more convincing than the original. To reduce the difference between the two distributions, we use CLIP Text-Image similarity to adversarially filter our splits. For each pristine sample $(img_1,cap_1)$ with CLIP Text-Image similarity $CTI_p$ we have two options: 
(1) There may exist a set of falsified candidates $(img_2,cap_1)$, where their score $CTI_f \geq CTI_p$, ordered for each of our splits: using (a) CLIP Text-Image, (b) CLIP Text-Text, (c) SBERT-WK Text-Text, (d) ResNet Place, respectively.
(2) There exists a set of candidates where their score $CTI_f < CTI_p$, ordered in the same way. We select the top scoring sample from set (1), else we select the top sample from (2) if set (1) is empty. Finally, we remove the sample with $max(CTI_p - CTI_f)$ until we get a 50-50 ratio of samples from sets (1) and (2) since the larger the delta $CTI_p - CTI_f$ the more likely the falsified sample is of low quality. As a result, on a ranking task where CLIP off-the-shelf is given a caption and two images, it correctly chooses the pristine image $50\%$ of the time by design.

\myparagraph{\DATASETNAME{} Statistics} The detailed statistics for the proposed \DATASETNAME{} are reported in Table~\ref{tab:dataset_stats}. Each caption appears twice, once in a pristine sample then again in a falsified sample. Thus exactly half of the samples are pristine and half are falsified, and there is no unimodal text bias in the dataset. We report the total number of samples across splits \emph{including any duplicates} as Total/Sum, and the number of \emph{unique} text-image pairings as Total/Unique in Table \ref{tab:dataset_stats}.

\begin{table}[htb]
\caption{\DATASETNAME{} Statistics.}
\label{tab:dataset_stats}
\vspace{-0.4cm}
\begin{center}
\begin{scriptsize}
\begin{tabular}{@{}l@{\ \ \ }|@{\ \ \ }r@{\ \ \ \ }r@{\ \ \ \ }r@{}}
\toprule
Split  & Train & Val & Test\\
\midrule
(a) Semantics/CLIP Text-Image & 453,128& 47,248& 47,288\\
(b) Semantics/CLIP Text-Text & 516,072 & 53,876& 54,164\\
(c) Person/SBERT-WK Text-Text & 17,768 & 1,756 & 1,816\\
(d) Scene/ResNet Place & 124,860 & 13,588 & 13,636 \\
\midrule
Total/Sum & 1,111,828 & 116,468 & 116,904 \\
Total/Unique & 816,922 & 85,609 & 85,752 \\
Merged/Balanced & 71,072 & 7,024 & 7,264 \\
\bottomrule
\end{tabular}
\end{scriptsize}
\end{center}
\vspace{-0.4cm}
\end{table}

Table~\ref{tab:dataset-comparison} provides a comparison to the most related prior datasets, highlighting the key differences, such as the image-text mismatch procedure used in each dataset.

\begin{table}[htb]
\caption{Comparison to prior related datasets. Size is the total number of unique samples across all splits.}
\label{tab:dataset-comparison}
\vspace{-0.4cm}
\begin{center}
\begin{scriptsize}
\begin{tabular}{@{}l@{\ }l@{\ \ }l@{\ \ }l@{\ \ }r@{}}
\toprule
Dataset & Data & Source  & Mismatch & Size\\
\midrule
MAIM & Caption, & Flickr & Random & 239k\\
~\cite{jaiswal2017multimedia} & Image & &\\
MEIR & Caption, & Flickr & Text entity  & 57k\\
~\cite{sabir2018deep} & Image, GPS & & manipulation \\
TamperedNews & Article, & BreakingNews & Text entity & 776k \\
~\cite{muller2020multimodal} & Image & & manipulation \\
COSMOS & Caption, & News Outlets & Two sources  & 453k\\
~\cite{aneja2021catching}& Image & &  (3k labeled) & \\
\midrule
\DATASETABBR{} (Ours) & Caption,  & VisualNews & Automatic  & 988k\\
 & Image &  &  retrieval & \\
\bottomrule
\end{tabular}
\end{scriptsize}
\end{center}
\vspace{-0.5cm}
\end{table}

\myparagraph{Dataset Examples} Here, we provide a few samples from the \DATASETNAME{}. Figure~\ref{fig:supplemental-comparison} compares the matches from each split for the same query caption and pristine image. Our diverse methods of computing similarity result in different weightings for concepts, displaying the realm of plausible images for a given caption. In (1), CLIP Text-Image matches ``parliament'' to Tennessee's governor speaking to a General Assembly, CLIP Text-Text matches ``Angela Merkel'' to Ingeborg Berggreen-Merkel speaking, and SBERT-WK Text-Text finds a match of Angela Merkel at a summit. In (2), CLIP Text-Image matches ``tsunami'' to a flooding in New York, CLIP Text-Text matches ``Japan'' to the president of a Japanese company, and ResNet Place matches ``earthquake'' to a destroyed highway after an earthquake in Chile\footnote{The Person split and Scene splits have no shared pristine samples since all matches either do or do not have `PERSON'' named entities depending on the split.}.

%% file: sections/experiments.tex
\section{\label{sec:experiments}Experiments}

We start by describing our experimental setup and then present the results of our benchmarking study.

\subsection{\label{sec:experimental_setup}Experimental Setup}

\myparagraph{Model Architectures}
For our base models we rely on CLIP~\cite{radford2learning} and VisualBERT~\cite{li2019visualbert}. We include VisualBERT as it is a representative recent model and is an appropriate baseline for addressing the semantic mismatch tasks.

\emph{CLIP} passes image and text through separate encoders that are trained to generate similar representations for related concepts. The model is pretrained on a web-based corpus of 400M image-text pairs using a contrastive loss, in which the cosine similarity of true image-text pairs is maximized. 

\emph{VisualBERT} passes image and text through a shared series of transformer layers to align them into one embedding space. For its bounding box features, we use a Faster-RCNN model ~\cite{ren2015faster} trained on Visual Genome with a ResNeXT-152 backbone. For pretraining, we only use the Masked Token Loss reported by ~\citet{li2019visualbert}, which masks each text token with probability 0.15.
We pretrain VisualBERT either on the 3M image-caption pairs from Conceptual Captions~\cite{sharma-etal-2018-conceptual}, based on alt-texts from web images stripped of all named entities, or on the 1M pairs from the VisualNews~\cite{liu2020visualnews}, based on captions from the news images. 

\myparagraph{Implementation Details} Our task is to \emph{classify} each image-caption pair as pristine or falsified. We fine-tune both models as we train the classifiers. When finetuning, we use a learning rate of 5e-5 for the classifier and 5e-7 for other layers. We train with a batch size of 32 for 88k steps for the Semantics splits and 44k steps for the Person and Scene splits. We report \emph{classification accuracy} over all samples (All) and separately for the Pristine and Falsified samples. We also report model performance at varying false alarm rates via ROC curves.

\subsection{\label{sec:experimental_results}Experimental Results}
In this section, we benchmark several methods on our proposed dataset to assess its difficulty. First, we compare the performance of unimodal vs. multimodal models to ensure that methods cannot exploit unimodal biases. Next, since we leverage CLIP ViT/B-32 to make our dataset challenging, we explore whether our task could be solved by a different model specifically pretrained on the news domain, leveraging a different backbone, or with more model parameters. In our final experiment, we train a single model on the union of all splits (Total/Sum in Table~\ref{sec:dataset}), while all the other experiments report the performance of the \textit{distinct models trained on each split individually}. All tables in this section evaluate on the same test set per split.

\begin{table*}[t]
\caption{Classification performance on the test set for the following models: (I) Image-only CLIP (w/ ViT-B/32), (II) Multimodal CLIP (w/ ViT-B/32), (III) VisualBERT-CC pretrained on the Conceptual Captions dataset, (IV) VisualBERT-VN pretrained on the Visual News.}
\vspace{-0.3cm}
\label{tab:results_merged}
\begin{center}
\begin{small}
\begin{tabular}{@{}l@{\ }|@{\ }c@{\ }|@{\ }c@{\ \ }c@{\ \ }c@{\ }|c@{\ }|@{\ }c@{\ \ }c@{\ \ }c@{}}
\toprule
& (I) & \multicolumn{3}{c|}{(II)} & (III) & \multicolumn{3}{c}{(IV)} \\
& CLIP Image-Only & \multicolumn{3}{c|}{CLIP} & VisualBERT-CC & \multicolumn{3}{c}{VisualBERT-VN} \\
\midrule
Split & All & All & Pristine & Falsified & All & All & Pristine & Falsified \\
\midrule
(a) Semantics/CLIP Text-Image & 0.5471 & 0.6698 & 0.7543 & 0.5853 & 0.5413 & 0.5774 & 0.6770 & 0.4778\\
(b) Semantics/CLIP Text-Text & 0.5247 &  0.6939 & 0.7409 & 0.6469 & 0.5714 & 0.5949 & 0.6591 & 0.5307\\
(c) Person/SBERT-WK Text-Text & 0.5000 & 0.6101 & 0.6178 & 0.6024 & {0.5947} & {0.6333} & 0.7247 & 0.5419\\
(d) Scene/ResNet Place & 0.5391 & 0.6821 & 0.7835 & 0.5807 & 0.5636 & 0.6112 & 0.6693 & 0.5532\\
Merged/Balanced  & 0.5288 & 0.6023 & 0.7007 & 0.5039 & 0.5482 & 0.5863 & 0.7841 & 0.3885\\
\bottomrule
\end{tabular}
\end{small}
\end{center}
\vspace{-0.2cm}
\end{table*}

\myparagraph{Unimodal Model Performance} One motivation for this work is that several prior works rely on automatic text manipulation to generate mismatched media. We argue that entity manipulation can introduce linguistic biases. We trained a text-only BERT model~\cite{devlin2019bert} on \emph{just the named entities} of the TamperedNews dataset (rather than the full articles) and achieved comparable results to the original paper's image-and-text based system \cite{muller2020multimodal}. For their ``Document Verification'' task, where the goal is to select one out of two articles given an image, we were able to achieve 90\% versus their 93\% on the Persons Country Gender (PsCG) split. For the Outdoor Places City Region split, GCD(25, 200), we were able to achieve 96\% versus their 76\%. %
This suggests that text manipulation can introduce biases that make the use of images unnecessary.

To avoid unimodal biases, our dataset is \emph{balanced} with respect to its captions (every caption is used once in a pristine sample and again in a falsified sample). Since we do not have such constraint on our images, we ran an image-only CLIP model (i.e. zeroing out the text inputs to CLIP) to verify that there is minimal visual bias. Based on our findings and due to the smaller size of the Person split (c), we additionally balance this particular split with respect to images, which means any image-only model is expected to achieve exactly 50\% accuracy on this split. As shown in Table ~\ref{tab:results_merged} (I), overall the image-only CLIP model obtains slightly above chance performance, significantly lower than the full image-text model, Table ~\ref{tab:results_merged} (II).

\myparagraph{Multimodal Model Performance} 
We report results for the multimodal CLIP-based classifiers in Table~\ref{tab:results_merged} (II). (Again, we repeat that here we train distinct classifiers for each split individually.) CLIP tends to ``over-predict'' pristine labels, indicating that many falsified samples are highly realistic and plausible. The Person split appears the most challenging, which could be partly explained by having the least number of samples. The Merged split, which contains an equal proportion of all four splits, is as difficult as its most difficult sub-split, seen by how CLIP classifies correctly $60\%$ of the time compared with $61\%$ for the Person split. 

On the other hand, VisualBERT-CC (pretrained on Conceptual Captions) in Table~\ref{tab:results_merged} (III) performs the best on the Person split with a performance of 59\% that approaches CLIP's. This indicates that the Person split primarily requires semantic understanding, and that a model with no knowledge of named entities can compete with a model that is strong at recognizing celebrities and other named entities. As expected, on all other splits that test entity understanding VisualBERT-CC performs on average 10\% worse than CLIP.

\myparagraph{Pretraining VisualBERT on News Domain} We also compare the performance of VisualBERT pretrained on Conceptual Captions (VisualBERT-CC) vs. VisualNews (VisualBERT-VN). In Table ~\ref{tab:results_merged} (III vs. IV) we observe that in-domain data, including named entities, provides a 3-5\% boost uniformly across all splits. Even more, with a training corpus less than 1\% the size of CLIP's training data, VisualBERT-VN is able to exceed CLIP performance on the Person split and approach CLIP performance on the Merged split. In fact, the largest gap between VisualBERT-VN and CLIP remains in the Semantics splits, where named entity understanding is crucial. Hence, through these results we can observe that VisualBERT-VN is strongest at semantic reasoning while CLIP is strongest at named entity recognition, which makes sense given their architectures (more deeply interactive VisualBERT-VN vs. more shallow CLIP).%

\begin{figure}[h]
\begin{center}
\includegraphics[width=\linewidth]{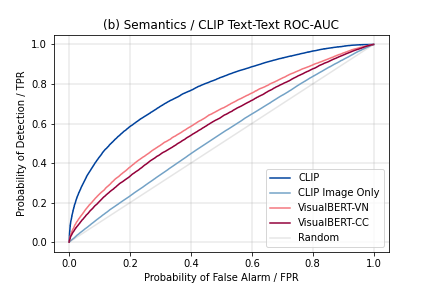}
\caption{Semantics/CLIP Text-Text ROC Curve}
\label{fig:roc_semantics_clip_text_text}
\end{center}
\end{figure}
\begin{figure}[h]
\begin{center}
\includegraphics[width=\linewidth]{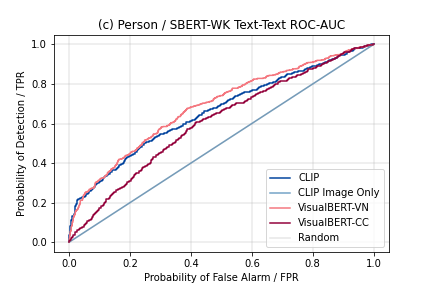}
\caption{Person/SBERT-WK Text-Text ROC Curve}
\label{fig:roc_person_sbert_text_text}
\end{center}
\end{figure}

\myparagraph{ROC Curves} We also include the ROC curves for the softmaxed logits produced by the models, see Figure \ref{fig:roc_semantics_clip_text_text}, \ref{fig:roc_person_sbert_text_text}. We see that the trends for these curves are consistent with the model rankings recorded in Table ~\ref{tab:results_merged}, with CLIP outperforming the other models by a wide margin on Semantics/CLIP Text-Text in Figure \ref{fig:roc_semantics_clip_text_text} across all false alarm rates. For Person/SBERT-WK Text-Text in Figure \ref{fig:roc_person_sbert_text_text}, VisualBERT-VN has virtually identical performance as CLIP at low false alarm rates. However, for systems that can tolerate more false alarms VisualBERT-VN shows a small advantage.

\myparagraph{Comparing CLIP Models} Recall that we used CLIP ViT-B/32 to construct our dataset. Here, we investigate whether our dataset could be ``solved'' by an existing CLIP model with a different backbone (ViT-B/32 vs. RN50) or a bigger model (RN50 vs. RN101). %
In Table ~\ref{tab:clip_accuracy_backbones}, we observe that RN50 performs slightly better than ViT-B/32 across the board, with at most 2\% performance difference between the two. We also see that while RN101 has more parameters than RN50, it only provides a small 1-2\% boost on most splits. The split where RN101 achieves the largest improvement (4\%) is the Merged/Balanced split, which aligns with the need for a model to capture more complex patterns to classify samples from multiple generation methods. Although we used a specific CLIP model architecture during dataset generation, we see that our dataset is still challenging for models with different architectures and more parameters.

\begin{table}[htb]
\caption{Comparing different CLIP backbones, classification performance (test set).}
\label{tab:clip_accuracy_backbones}
\vspace{-0.3cm}
\begin{center}
\begin{small}
\begin{tabular}{@{}l@{\ }|c@{\ \ }|c@{\ \ \ }c@{\ \ }c@{}}
\toprule
Split & Model & All & Pristine & Falsified\\
\midrule
(a) Sem/CLIP T-I  & ViT-B/32 & 0.6698 & 0.7543 & 0.5853\\
    & \textbf{RN50} & \textbf{0.6824} & \textbf{0.7461} & \textbf{0.6188}\\
    & RN101 & 0.6765 & 0.7444 & 0.6085\\
(b) Sem/CLIP T-T & ViT-B/32 & 0.6939 & 0.7409 & 0.6469\\
    & RN50 & 0.7182 & 0.7486 & 0.6878\\
    &\textbf{RN101} &\textbf{0.7244} & \textbf{0.7442} & \textbf{0.7046}\\
(c) Per/SB-WK T-T & ViT-B/32 & 0.6101 & 0.6178& 0.6024\\
    & RN50 & 0.6123 & 0.7357 & 0.4890\\
    &\textbf{RN101} &\textbf{0.6393} & \textbf{0.7004} & \textbf{0.5782}\\
(d) Scene/RN Place & ViT-B/32 & 0.6821 & 0.7835 & 0.5807\\
    & RN50 & 0.7004 & 0.7765 & 0.6244\\
    &\textbf{RN101} &\textbf{0.7137} & \textbf{0.7712} & \textbf{0.6562}\\
Merged/Balanced & ViT-B/32 & 0.6023 & 0.7007 & 0.5039\\
    & RN50 & 0.6162 & 0.6836 & 0.5487\\
    &\textbf{RN101} &\textbf{0.6597} & \textbf{0.6768} & \textbf{0.6426}\\
\bottomrule
\end{tabular}
\end{small}
\end{center}
\vskip -0.1in
\end{table}

\myparagraph{Evaluating A Single Unified Model}
Finally, we explore whether it is beneficial to combine various splits during training. Unlike Tables~\ref{tab:results_merged}, \ref{tab:clip_accuracy_backbones} which evaluate \emph{separate} models trained on each individual split, here we evaluate a \emph{single} model trained on all the splits jointly, see Table~\ref{tab:clip_total_accuracy}. The Total/Sum set (introduced in Table~\ref{tab:dataset_stats}) combines the samples from all the splits, so that it is balanced with respect to pristine and falsified labels but has different proportions of each type, e.g. around 87\% of samples are from the Semantics splits (a,b).

\begin{table}[htb]
\caption{CLIP (ViT/B-32) test set classification performance when training a single model with all the available training samples, i.e. Total / Sum in Table~\ref{tab:dataset_stats}.}
\label{tab:clip_total_accuracy}
\vspace{-0.3cm}
\begin{center}
\begin{small}
\begin{tabular}{@{}l@{\ \ }|@{\ \ }c@{\ \ }c@{\ \ }c@{}}
\toprule
Split & All & Pristine & Falsified\\
\midrule
(a) Semantics/CLIP Text-Image & 0.6651 & 0.7582 & 0.5720 \\
(b) Semantics/CLIP Text-Text & 0.6457 & 0.7563 &  0.5351\\
(c) Person/SBERT-WK Text-Text & 0.6399 & 0.7434 & 0.5363\\
(d) Scene/ResNet Place & 0.6824 & 0.7778 & 0.5870 \\
Merged/Balanced & 0.6611 & 0.7574 & 0.5647\\
\bottomrule
\end{tabular}
\end{small}
\end{center}
\vskip -0.1in
\end{table}

Comparing Table ~\ref{tab:results_merged} (II) with Table ~\ref{tab:clip_total_accuracy}, we note that the Person split experiences a 2\% boost in performance even though it represents only 1\% of the training data. Clearly, it benefits from the other sample types. We also note the 5\% degradation in performance for the Semantics/CLIP Text-Text, likely due to the challenges in learning to address several mismatch types at once\footnote{We hypothesize that this may be due to the joint training with the Person samples -- if a model does not know who the pictured individual is, then the mismatches in Semantics/CLIP Text-Text may look similar to those in the Person split, as they both are matched using only textual information.}. Finally, we see a boost of almost 6\% for the Merged/Balanced set, showing the benefit of training in a unified setting for this more realistic split. One other trend we notice is that the Pristine accuracy seems to overall benefit more than the Falsified accuracy.

%% file: sections/additional.tex
\section{\label{sec:additional} Additional Analysis}
In this section, we gain further insights into the quality of our dataset via human evaluation and saliency map analysis. With the human evaluation, we assess whether our dataset could fool humans and pose a realistic threat. We also assess whether our dataset may have ``unsolvable'' true matches that in fact do not misrepresent anything. With our qualitative saliency map analysis, we investigate if the automatic models are learning to leverage high level semantic or entity cues after training on our dataset.

\begin{table}[h]
\caption{Human Performance on 200-sample subset of Merged/Balanced. ``Optimistic'' accuracy is defined as at least 1 worker gave the correct answer.}
\label{tab:human-eval}
\begin{center}
\begin{small}
\begin{tabular}{l|ccr}
\toprule
  & All & Pristine & Falsified\\
\midrule
Average & 0.656 & 0.962 & 0.350 \\
Optimistic & 0.845 & 1.000 & 0.690 \\
\bottomrule
\end{tabular}
\end{small}
\end{center}
\end{table}

\myparagraph{Human Performance}
Here, we estimate the difficulty of the proposed task for \emph{humans}, aiming to assess how convincing our automatically matched images and captions are. We randomly select a set of 200 samples from the Merged/Balanced split, with an equal number of samples from all types (50 from each split, where 25 are pristine and 25 are falsified). We conduct our evaluation on Amazon Mechanical Turk\footnote{www.mturk.com}. For each image-caption pair we ask $5$ workers the following three questions: (a) ``Could this image belong to the given caption?'' (Yes/No), (b) ``How confident are you in your answer?'' (1: Very, 2: Somewhat, 3: Not at all), (c) ``Would it help to use a search engine to be more confident?'' (Yes/No). Note, that we specifically instruct the workers \textbf{not} to use search engines, to prevent them from discovering the original news articles on the Web. The key takeaways from the evaluation are as follows. (1) The average accuracy over all samples is $0.656$, while the most ``optimistic'' accuracy (at least 1 worker gave the correct answer) is $0.845$. This clearly shows that the task is not easy for humans. For reference, our CLIP model trained on the Total/Sum set (Table~\ref{tab:clip_total_accuracy}) achieves $0.6650$ on these 200 samples, essentially \emph{matching human performance}. (2) Humans are much better in recognizing pristine than falsified samples, with an average accuracy of $0.962$ and $0.350$ respectively. This shows that they are often misled by our falsified matches. %
(3) The ``optimistic'' accuracy for falsified samples is $0.690$, meaning that majority are still solvable with just the prior knowledge of those workers. Among the 31 samples that all workers classified incorrectly, we estimate that $67\%$ are answerable with additional knowledge of person identity and other context cues.
(4) While the average confidence score is $1.755$, the confidence on the correctly vs. incorrectly predicted samples is $1.658$ and $1.940$ respectively (lower is better), i.e. humans were more confident on samples they predicted correctly. 
(5) The average accuracy when there is a reported ``need to use a search engine'' is $0.589$ vs. $0.760$ otherwise. This shows that the humans do better when they encounter familiar concepts vs. less familiar ones. Hence, additional search is likely to boost the results, as we have observed in our own internal analysis. (6) Across the four types of mismatch, the easiest for humans is Scene, followed by Semantics/CLIP Text-Text, Semantics/CLIP Text-Image, and finally the Person split. Interestingly, this overall aligns with the trends observed for the automatic methods.

\myparagraph{Qualitative Analysis}
Finally, we analyze CLIP ViT saliency maps and prediction using the method presented in ~\citet{chefer2021generic}. We select examples from the 200 samples used in our human evaluation.
\begin{figure}[t]
\begin{center}
\includegraphics[width=\linewidth]{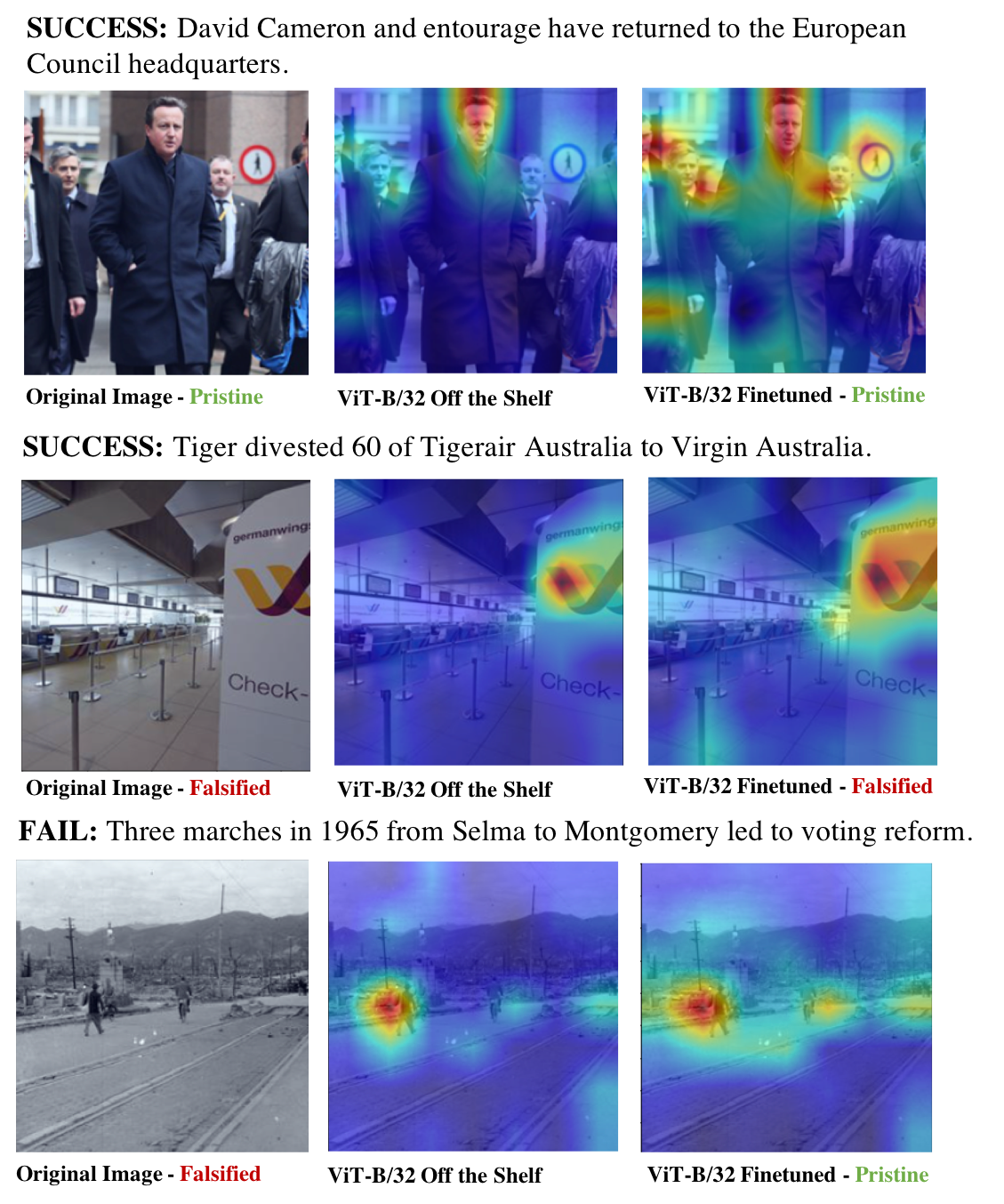}
\vspace{-0.5cm}
\caption{Qualitative examples of CLIP ViT-B/32 success and failure cases with salience visualization.}
\label{fig:clip-success-fail}
\vspace{-0.5cm}
\end{center}
\end{figure}
As seen in Figure ~\ref{fig:clip-success-fail}, finetuning on our dataset often forces CLIP to focus on salient objects mentioned in the caption beyond the person of interest, for example expanding its attention from David Cameron to ``entourage.`` CLIP also has a number of capabilities off-the-shelf that require minimal finetuning, for example sign reading and logo recognition in the case of distinguishing ``Tigerair Australia'' from ``Germanwings.'' We also present a failure case where CLIP focuses on the two people in the foreground when the caption talks about ``marches.`` Evidently the model does find one point of support -- the photo looks like it could have been taken in 1965 -- but it fails to identify the absence of a crowd which would have been a ``red flag'' for a human. Similar failure cases, such as a falsified caption that mentions Mitt Romney and Rand Paul at a rally but only pictures Romney, highlight how our dataset can be particularly challenging because pristine news is an ambiguous domain that often mentions entities but does not picture them.

%% file: sections/conclusion.tex
\section{\label{sec:conclusion}Conclusion}

We introduced \textbf{\DATASETABBR{}}, a large-scale automatically constructed dataset for classifying news image-caption pairs as real or out-of-context. We found CLIP to be effective for the dataset construction and recognition of mismatches. By design, unimodal models cannot solve our task, while multimodal ones require named entity and semantic knowledge to do well on our diagnostic splits. Our Merged set aims to model the more realistic diversity of image-caption mismatches in the wild.

From our experimental results, we find that the ResNet backbone offers a modest performance boost compared to a ViT-B/32 model. We find that a CLIP ViT model is able to match human performance on a small subset of our Merged / Balanced split, and that our task is generally difficult with an average 66\% human accuracy. 

Our training data could be used to augment and increase the training data size of human-made falsified news, which often lack ground truth labels or sufficient scale. Overall, we show that it is possible to automatically match plausible images for given input captions, and we present a challenging benchmark to foster the development of defenses against large-scale image repurposing.

%% file: sections/ethics.tex
\section{\label{sec:ethics}Ethical Considerations}
Here, we discuss ethical considerations regarding our dataset. Image repurposing is a prominent societal issue that lacks sufficient training data, as both human generation and annotation are costly. Even more, our work aims to be proactive in characterizing this new threat of machine generated misinformation and proposes a number of solutions that serve as baselines for detection. By presenting a number of techniques and key observations about falsified out-of-context news, we hope that our dataset serves as a net benefit for society. 

\boldparagraph{How was the data collected?} Our dataset was automatically generated using features from CLIP ~\cite{radford2learning}, SBERT-WK ~\cite{wang2020sbert}, a ResNet50 trained on Places365 \cite{zhou2017places}, and other additional metadata. Our dataset is composed of intelligent automatic matches for the VisualNews~\cite{liu2020visualnews} images and captions that occur in the real world.

\boldparagraph{What are the intellectual property rights?} The copyright and usage rights of our dataset are subject to that of VisualNews ~\cite{liu2020visualnews}. Our key contribution is an automatic dataset generation \textit{approach} that can be applied to any news dataset.

\boldparagraph{How did we address participant privacy rights?} N/A 

\boldparagraph{Were annnotators treated fairly? Did we require review from a review board?} N/A

\boldparagraph{Which populations do we expect our dataset to work for?} As our dataset is composed of news from The Guardian, BBC, USA Today, and The Washington Post, it largely focuses on events and people from Western countries like the US and the UK in addition to world news.

\boldparagraph{What is the generalizability of our claims?} We expect our results regarding the dangers of automatic image repurposing and experimental results of model detection performance to primarily apply to Western news in the English language.

\boldparagraph{How did we ensure dataset quality?} We made a number of design choices noted in Sections~\ref{sec:dataset},\ref{sec:experiments} to remove any unimodal biases and  ensure that samples would be challenging for recent AI models. We also conducted a human evaluation and found that our matches were challenging for humans.

\boldparagraph{What is the climate impact?} 
For dataset construction, excluding the fixed cost of embedding extraction, matching takes about 1-2 hours with 8 GPUs to process 400k pristine samples. This is an estimated 1.28 kg CO$_2$ eq emissions. Each finetuning experiment takes about 9 hours on one GPU, which is an estimated 0.97 kg CO$_2$ eq emissions. Estimations were conducted using the \href{https://mlco2.github.io/impact#compute}{MachineLearning Impact calculator} presented in \citet{lacoste2019quantifying}.

\boldparagraph{What are the potential dataset biases?}
Here, we discuss the models used to compute matching scores during dataset generation to understand the potential biases that may appear in our dataset. 
\textbf{CLIP:} ~\citet{radford2learning} conducts a number of experiments on race %
and gender %
prediction to investigate potential biases learned from their large-scale web corpus. They report that CLIP had significant disparities when classifying individuals from different races into crime-related and non-human categories. They also report gender differences when attaching appearance-related terms and occupations to photos of Members of Congress. \\
\textbf{SBERT-WK:} Since SBERT-WK dissects the embedding of BERT~\cite{devlin2019bert} to compute semantic similarity, any biases present in the model are from those learned by BERT. Prior works note that BERT does encode a number of social biases, including a strong association between gender and career/family or math/arts ~\cite{kurita-etal-2019-measuring}. \\
\textbf{Places365:} From our qualitative analysis, we found that the Places365 dataset exhibited race and age associations with certain scene labels. As a result, we only used the ResNet embeddings when computing matches and completely ignored the labels to somewhat mitigate these biases.

\boldparagraph{How might this work contribute to the spread of disinformation?} We acknowledge that our work can be misused to mass-generate repurposed images. Adversaries can now automate both synthesizing an inflammatory news piece using recent models like GPT-3 \cite{brown2020language} or Grover \cite{zellers2019defending}, and they can retrieve an appropriate image using models we present such as CLIP \cite{radford2learning}, SBERT-WK \cite{wang2020sbert}, and ResNet \cite{zhou2017places}. %

However, we argue that our work cannot be immediately used to generate targeted attacks. An adversary would either have to generate synthetic captions (since our models are trained on human-made captions, this would mean a domain shift) or manually write captions (which is time and money consuming) tailored for their narrative. Although we demonstrate that automatic image repurposing can be convincing to humans, additional effort is required to produce \textit{malicious} image-caption pairs.

%% file: sections/appendix.tex
\algnewcommand\algorithmicforeach{\textbf{for each}}
\algdef{S}[FOR]{ForEach}[1]{\algorithmicforeach\ #1\ \algorithmicdo}
\algnewcommand{\LineComment}[1]{\State \(\triangleright\) #1}

\appendix
\section{Appendix to ``NewsCLIPpings: Automatic Generation of Out-of-Context Multimodal Media''}

In the following we include additional details regarding dataset construction (Section~\ref{sec:supp_dataset_construction}), additional experimental results (Section~\ref{sec:supp_results}), and dataset examples (Section~\ref{sec:supp_dataset}).

\subsection{Dataset Construction}
\label{sec:supp_dataset_construction}
Here, we go over additional implementation details for dataset construction. (1) To ensure the quality of our dataset, we filter pristine samples such that they have between 5 and 30 words, at least 2 named entities, and non-corrupted images.
(2) We split our 509,730 pristine image-caption pairs into chunks of size $\sim$40k for train/val/test. We only computed matches across these disjoint chunks, which allowed us to parallelize the generation process. (3) We precomputed features such as spaCy NER, Radboud Entity Linking, SBERT-WK text embeddings, CLIP text embeddings, CLIP image embeddings, Faster R-CNN bounding boxes, ResNet50 place embeddings. (4) We ran Algorithm ~\ref{alg:matching_algorithm}, the {\DATASETABBR} matching algorithm. (5) We removed low quality samples by balancing the number of samples where the CLIP text-image score is higher for the pristine vs falsified pair as described in our Adversarial CLIP Filtering process in Section ~\ref{sec:dataset}.

\begin{algorithm}
\caption{{\DATASETABBR}  Matching}\label{alg:matching_algorithm}
\textbf{Input:} Dataset $D$, scoring function $sim \in$ \{CLIP Text-Image, CLIP Text-Text, SBERT-WK Text-Text, ResNet Place\}, split type $split$.\\
\textbf{Output:} Matches $M$ \\
\begin{algorithmic}
\State $M \gets \{\}$
\ForEach{$p =(i_1, c_1) \in D$}
    \State $M_H, M_L \gets \{\}, \{\}$
    \State $D* \gets$ sort($D$, key=$sim$)
    \ForEach{$(i_2, c_2) \in D^*$}
    \State $f \gets (i_2, c_1)$
    \If{filter($p, f, split$)} \\
        \quad\quad\quad\quad \textbf{continue}
    \ElsIf{$CTI(f) > CTI(p)$}
        \State $M_H \gets M_H \cup \{f\}$
    \Else
        \State $M_L \gets M_L \cup \{f\}$
    \EndIf
    \EndFor
    \State $M \gets M \cup  \{(M_H \cup M_L)[0]\}$
    \State $M \gets M \cup  \{p\}$
\EndFor \\
\end{algorithmic}
\end{algorithm}

\subsection{Additional Results}
\label{sec:supp_results}

\myparagraph{ROC Curves}
We include ROC curves for the splits not depicted in Section ~\ref{sec:experiments} in the main paper. Note that the rankings for model performance across all these splits are consistent with Table ~\ref{tab:results_merged}. We see that CLIP vastly outperforms the other models in the Semantics/CLIP Text-Image split whereas model performance is very similar in the Merged/Balanced split. This makes sense since CLIP was used as the scoring function for matches in Semantics/CLIP Text-Image, whereas a diverse set of scoring functions were used for Merged/Balanced.

\begin{figure*}[h]
\begin{center}
\includegraphics[width=\linewidth]{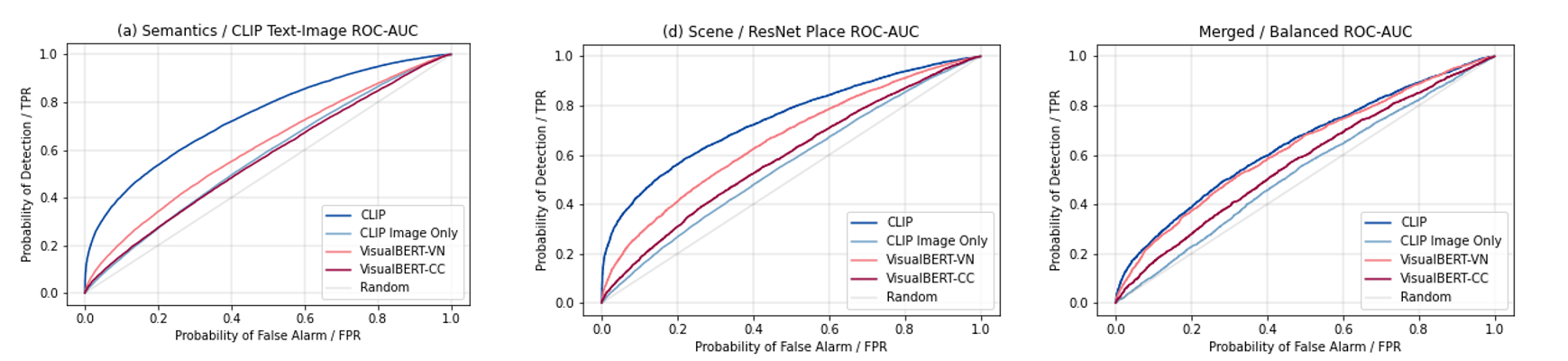}
\caption{Semantics/CLIP Text-Image, Scene/ResNet Place, Merged/Balanced ROC Curve}
\end{center}
\end{figure*}

\myparagraph{Finetuning CLIP Representation} We also experiment with freezing a varying number of CLIP layers to determine how useful CLIP's original knowledge and representation is for each split. We compare three models: RN50-all-frozen (RN50-af, no CLIP layers finetuned), RN50-lower-frozen (RN50-lf, final few layers finetuned) \footnote{The exact layers we finetune are ["visual.layer4", "visual.attnpool", "transformer.resblocks.11", "ln\_final", "text\_projection", "logit\_scale"].}, and RN50 (all layers finetuned). In Tables~\ref{tab:clip_accuracy_frozen_val} and \ref{tab:clip_accuracy_frozen}, we observe that finetuning all layers (RN50) benefits the Semantics splits (a,b) while freezing some or all layers (RN50-af/lf) benefits the Person and Scene splits (c,d). We posit that this result could be related to the training data size -- since the Semantics splits are significantly large we can meaningfully finetune all layers whereas in the others we do not have enough data to do so.

\begin{table}[h]
\caption{Comparing different finetuning strategies for CLIP, classification performance (val set).}
\label{tab:clip_accuracy_frozen_val}
\vspace{-0.5cm}
\begin{center}
\begin{small}
\begin{tabular}{@{}l@{\ \ }|@{\ \ \ }c@{\ \ \ }|c@{}}
\toprule
  & Model & All\\
\midrule
(a) Semantics/CLIP Text-Image  & RN50-af & 0.6354\\
    & RN50-lf & 0.6706\\
    & \textbf{RN50} & \textbf{0.6808}\\
(b) Semantics/CLIP Text-Text & RN50-af & 0.6826\\
    & RN50-lf & 0.6954\\
    & \textbf{RN50} &\textbf{0.7138}\\
(c) Person/SBERT-WK Text-Text & \textbf{RN50-af} & \textbf{0.6606}\\
    & RN50-lf & 0.6560\\
    & RN50 & 0.6167\\
(d) Scene/ResNet Place & RN50-af & 0.6965\\
    & \textbf{RN50-lf} & \textbf{0.7034}\\
    & RN50 & 0.6999\\
(e) Merged/Balanced & RN50-af & 0.6642\\
    & \textbf{RN50-lf} &\textbf{ 0.6684}\\
    & RN50 & 0.6468\\
\bottomrule
\end{tabular}
\end{small}
\end{center}
\vspace{-0.5cm}
\end{table}

\begin{table}[htb]
\caption{Comparing different finetuning strategies for CLIP, classification performance (test set).}
\label{tab:clip_accuracy_frozen}
\vspace{-0.5cm}
\begin{center}
\begin{small}
\begin{tabular}{@{}l@{\ \ }|c@{\ \ }|c@{\ \ \ }c@{\ \ }c@{}}
\toprule
Split  & Model & All & Pristine & Falsified\\
\midrule
(a) Sem/CLIP T-I  & RN50-af & 0.6372 & 0.6677 & 0.6068\\
    & RN50-lf &  0.6500 & 0.6840 & 0.6163\\
    & \textbf{RN50} & \textbf{0.6824} & \textbf{0.7461} & \textbf{0.6188}\\
(b) Sem/CLIP T-T & RN50-af & 0.6860 & 0.7167 & 0.6554\\
    & RN50-lf & 0.7005 & 0.7211 & 0.6800\\
    & \textbf{RN50} & \textbf{0.7182} & \textbf{0.7486} & \textbf{0.6878}\\
(c) Per/SB-WK T-T & \textbf{RN50-af} & \textbf{0.6669} & \textbf{0.6641} & \textbf{0.6696}\\
    & RN50-lf & 0.6547 & 0.6575 & 0.6520\\
    & RN50 & 0.6123 & 0.7357 & 0.4890\\
(d) Scene/RN Place & RN50-af & 0.6945 & 0.7543 & 0.6346\\
    & \textbf{RN50-lf} & \textbf{0.7028} &\textbf{0.7646} & \textbf{0.6411}\\
    & RN50 & 0.7004 & 0.7765 & 0.6244\\
Merged/Balanced & \textbf{RN50-af} & \textbf{0.6732} & \textbf{0.6726} & \textbf{0.6737}\\
    & RN50-lf & 0.6657 & 0.6952 & 0.6363\\
    & RN50 & 0.6162 & 0.6836 & 0.5487\\
\bottomrule
\end{tabular}
\end{small}
\end{center}
\vspace{-0.5cm}
\end{table}

\myparagraph{Ensembling Performance}
Next, we compare the predictions made by our best VisualBERT and best CLIP-based model (Table ~\ref{tab:vb_clip_overlap_union}). Specifically, we ask whether these very different models exhibit distinct behavior and have complementary skill sets. First, we assess how often both models are correct (Overlap) or at least one of them is correct (Union). As we see, the overlap tends to be near 40\%, while the ``optimistic'' union boosts accuracy to over 80\%. This clearly shows that the two models indeed have complementary strengths. We build a simple ensemble of VisualBERT and CLIP, in which we average the normalized logits of each model and classify accordingly (Avg). This scheme only gives a 1-2\% boost, which possibly indicates that CLIP is more ``opinionated'' and yields more disparate logits per class, forcing the ensemble to perform similarly to CLIP. This shows that leveraging both models may be promising but is non-trivial and should be explored by future work.

\begin{table}[htb]
\caption{Exploring complementarity of VisualBERT-VN and CLIP-RN101, reported overall classification performance (test set).}
\label{tab:vb_clip_overlap_union}
\vspace{-0.5cm}
\begin{center}
\begin{scriptsize}
\begin{tabular}{@{}l|@{\ \ }c@{\ \ }c@{\ \ }|@{\ \ }c@{\ \ }|c@{\ \ }c@{}}
\toprule
Split & VB-VN & CLIP-RN101 & Avg & Overlap & Union\\
\midrule
(a) Sem/CLIP T-I  & 0.5774 & \textbf{0.6765} & 0.6747 & 0.4341 & 0.8197\\
(b) Sem/CLIP T-T & 0.5949 & \textbf{0.7244} & 0.7234 & 0.4663 & 0.8530\\
(c) Per/SB-WK T-T  & 0.6333 & 0.6393 & \textbf{0.6553} & 0.4576 & 0.8150\\
(d) Scene/RN Place & 0.6112 & 0.7137 & \textbf{0.7230} & 0.4723 & 0.8527 \\
Merged/Balanced & 0.5863 & 0.6597 & \textbf{0.6662} & 0.4145 & 0.8315\\
\bottomrule
\end{tabular}
\end{scriptsize}
\end{center}
\vskip -0.1in
\end{table}

\myparagraph{Results on the Validation Set}
We include the results that correspond to Tables~\ref{tab:results_merged}, \ref{tab:clip_accuracy_backbones}, \ref{tab:clip_total_accuracy} of the main paper but evaluated on the validation set, see Tables~\ref{tab:results_merged_val}, \ref{tab:clip_accuracy_backbones_val}, \ref{tab:clip_total_accuracy_val}. Note, that the validation and test sets of our dataset were generated using identical techniques and should be comparable in composition. As a result, these tables demonstrate the same overall trends as reported in our main paper.

\begin{table*}[h]
\caption{Classification performance on the val set for the following models: (I) Image-only CLIP (w/ ViT-B/32), (II) Multimodal CLIP (w/ ViT-B/32), (III) VisualBERT-CC pretrained on the Conceptual Captions dataset, (IV) VisualBERT-VN pretrained on the Visual News.}
\label{tab:results_merged_val}
\begin{center}
\begin{small}
\begin{tabular}{@{}l@{\ }|@{\ }c@{\ }|@{\ }c@{\ \ }|c@{\ }|@{\ }c@{\ \ }}
\toprule
& (I) & (II) & (III) & (IV) \\
& CLIP Image-Only & CLIP & VisualBERT-CC & VisualBERT-VN \\
\midrule
Split & All & All & All & All \\
\midrule
(a) Semantics/CLIP Text-Image & 0.5472&  0.6711 & 0.5451 & 0.5773\\
(b) Semantics/CLIP Text-Text &0.5266 &  0.6921 & 0.5738 & 0.5963\\
(c) Person/SBERT-WK Text-Text & 0.5000& 0.6259 & 0.5666 & 0.6139\\
(d) Scene/ResNet Place & 0.5424 & 0.6803 & 0.5558 & 0.6128\\
Merged/Balanced & 0.5185& 0.6048 & 0.5504 & 0.5893\\
\bottomrule
\end{tabular}
\end{small}
\end{center}
\end{table*}

\begin{table}[h]
\caption{Comparing different CLIP backbones, classification performance (val set).}
\label{tab:clip_accuracy_backbones_val}
\begin{center}
\begin{small}
\begin{tabular}{@{}l@{\ \ }|@{\ \ \ }c@{\ \ \ }|c@{}}
\toprule
  & Model & All\\
\midrule
(a) Semantics/CLIP Text-Image  & ViT-B/32 & 0.6475\\
    & \textbf{RN50} & \textbf{0.6808}\\
    & RN101 & 0.6741\\
(b) Semantics/CLIP Text-Text & ViT-B/32 & 0.6921\\
    & RN50 & 0.7138\\
    &\textbf{RN101} &\textbf{0.7189}\\
(c) Person/SBERT-WK Text-Text & ViT-B/32 & 0.6099\\
    & RN50 & 0.6167\\
    &\textbf{RN101} &\textbf{0.6395}\\
(d) Scene/ResNet Place & ViT-B/32 & 0.6803\\
    & RN50 & 0.6999\\
    &\textbf{RN101} &\textbf{0.7153}\\
(e) Merged/Balanced & ViT-B/32 & 0.6048\\
    & RN50 & 0.6468\\
    &\textbf{RN101} &\textbf{0.6676}\\
\bottomrule
\end{tabular}
\end{small}
\end{center}
\end{table}

\begin{table}[h]
\caption{CLIP (ViT/B-32) val set classification performance when training a single model with all the available training samples, i.e. Total/Sum in Table~\ref{tab:dataset_stats}.}
\label{tab:clip_total_accuracy_val}
\begin{center}
\begin{small}
\begin{tabular}{@{}l@{\ \ }|@{\ \ }c@{\ \ }c@{\ \ }c@{}}
\toprule
Split & All\\
\midrule
(a) Semantics/CLIP Text-Image & 0.6632\\
(b) Semantics/CLIP Text-Text & 0.6445\\
(c) Person/SBERT-WK Text-Text & 0.6395\\
(d) Scene/ResNet Place & 0.6858\\
Merged/Balanced & 0.6640\\
\bottomrule
\end{tabular}
\end{small}
\end{center}
\end{table}

\subsection{Additional Dataset Details}
\label{sec:supp_dataset}

\begin{table}[t]
\begin{center}
\begin{scriptsize}
\begin{tabular}{@{}l|rrrrr@{}}
\toprule
Split  & (a)  & (b)  & (c)  & (d)\\
\midrule
(a) Semantics/CLIP Text-Image & 1.0000 & 0.1133 & 0.0000 &  0.0813\\
(b) Semantics/CLIP Text-Text & 0.1133 & 1.0000 & 0.0000 & 0.0888 \\
(c) Person /SBERT-WK Text-Text & 0.0000 & 0.0000 & 1.0000 & 0.0000\\
(d) Scene/Place Label & 0.0813 & 0.0888 & 0.0000 & 1.0000\\
\bottomrule
\end{tabular}
\end{scriptsize}
\end{center}
\caption{Ratio of exact overlap across splits.}
\label{tab:dataset-overlap}
\end{table}

Next, we report the amount of overlap across splits of our dataset (see Table~\ref{tab:dataset-overlap}) and find that they are relatively distinct, with a maximum of 11\% overlap in falsified matches for the two Semantics splits. 

Figures \ref{fig:supplemental-random-1} and \ref{fig:supplemental-random-2} depict \emph{randomly} selected samples from the train/val/test set of each respective split. Note how matches are highly plausible in our Semantics splits (a,b) but can be solved if the identity of a person is known or even with subtle semantic cues (e.g. an American flag when the caption describes a European person in the top left or that the image shows a letter when the caption describes a banner on the bottom left for Semantics/CLIP Text-Image). Also note, how in the Semantics/CLIP Text-Text, textual concepts from a query may impact the resulting retrieved image based on its own caption, e.g., ``flood'' in the top-right example. Note the challenging examples from our Person split (c), including some rather ambiguous ones such as top-right with Hillary Clinton, that require recognizing the context in which the person appears. Finally, in our Scene split (d) we show that the events in captions may be plausibly ``illustrated'' by purely leveraging visual scene similarity. Note again, that these are \emph{randomly} selected samples from our dataset.
\input{sections/examples}

%% file: sections/examples.tex
\label{sec:examples}

\onecolumn

\begin{figure}
\begin{center}
\includegraphics[width=16cm]{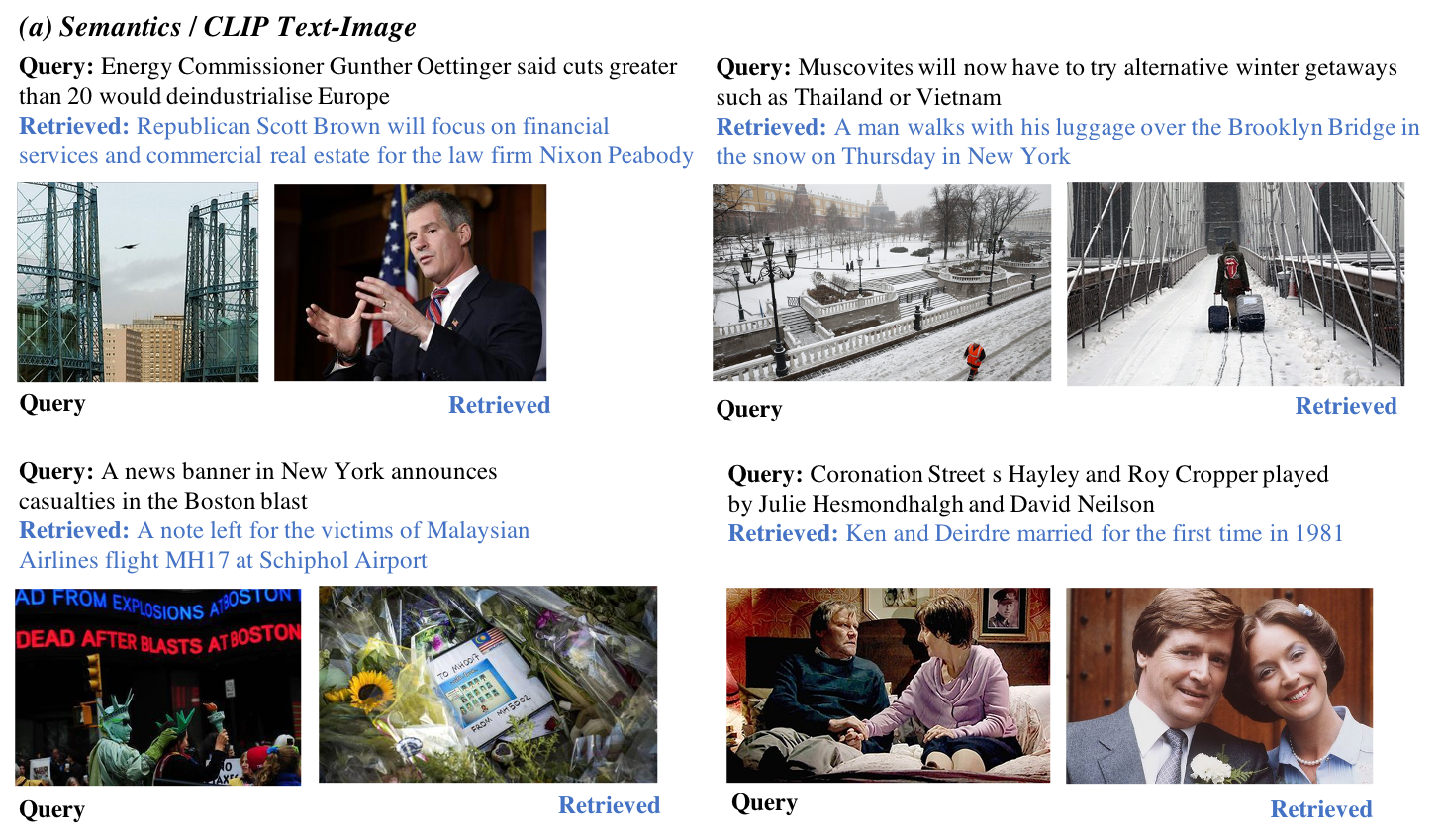}
\includegraphics[width=16cm]{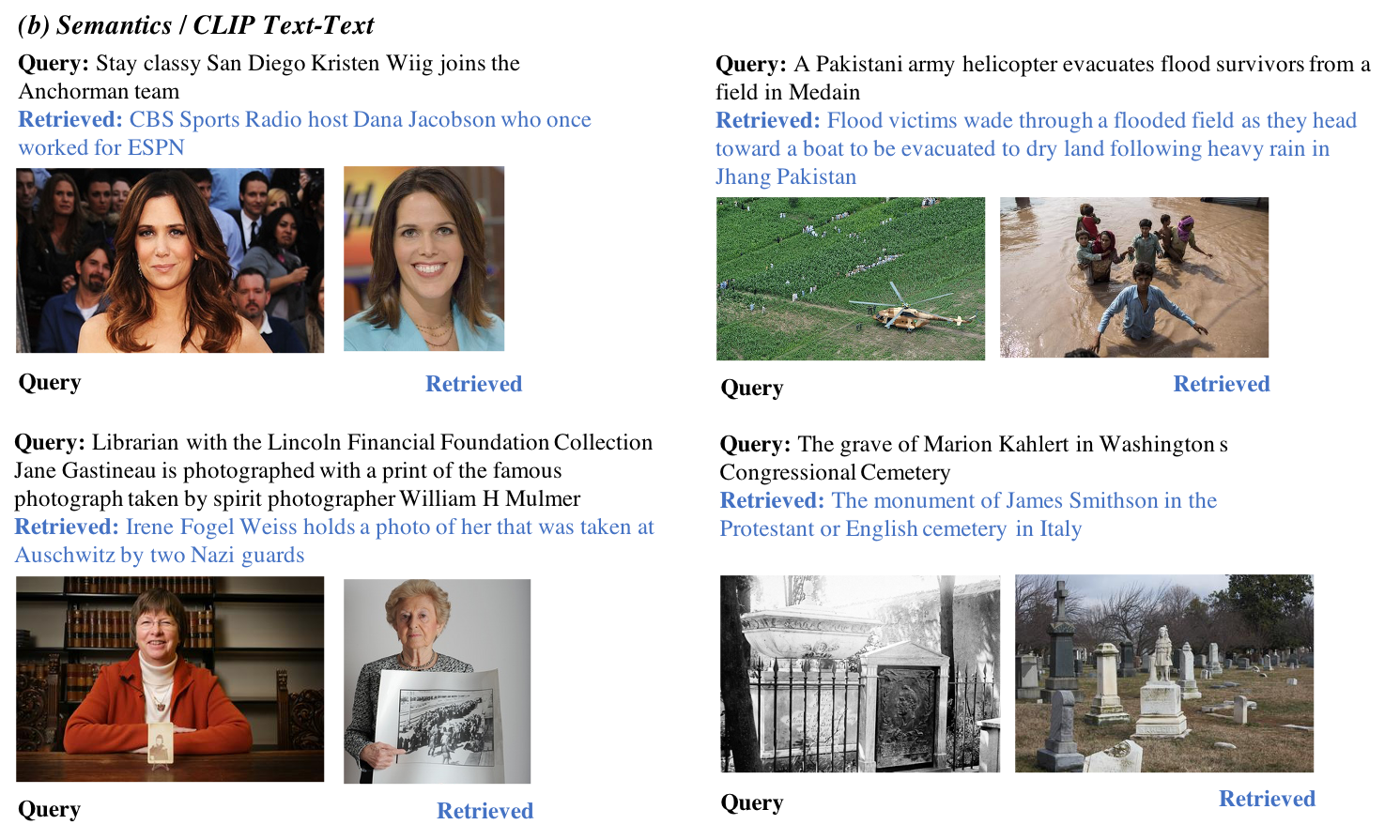}
\caption{Randomly selected (a), (b) samples from the train/val/test samples of each respective split.}
\label{fig:supplemental-random-1}
\vspace{-0.8cm}
\end{center}
\end{figure}

\begin{figure}
\begin{center}
\includegraphics[width=16cm]{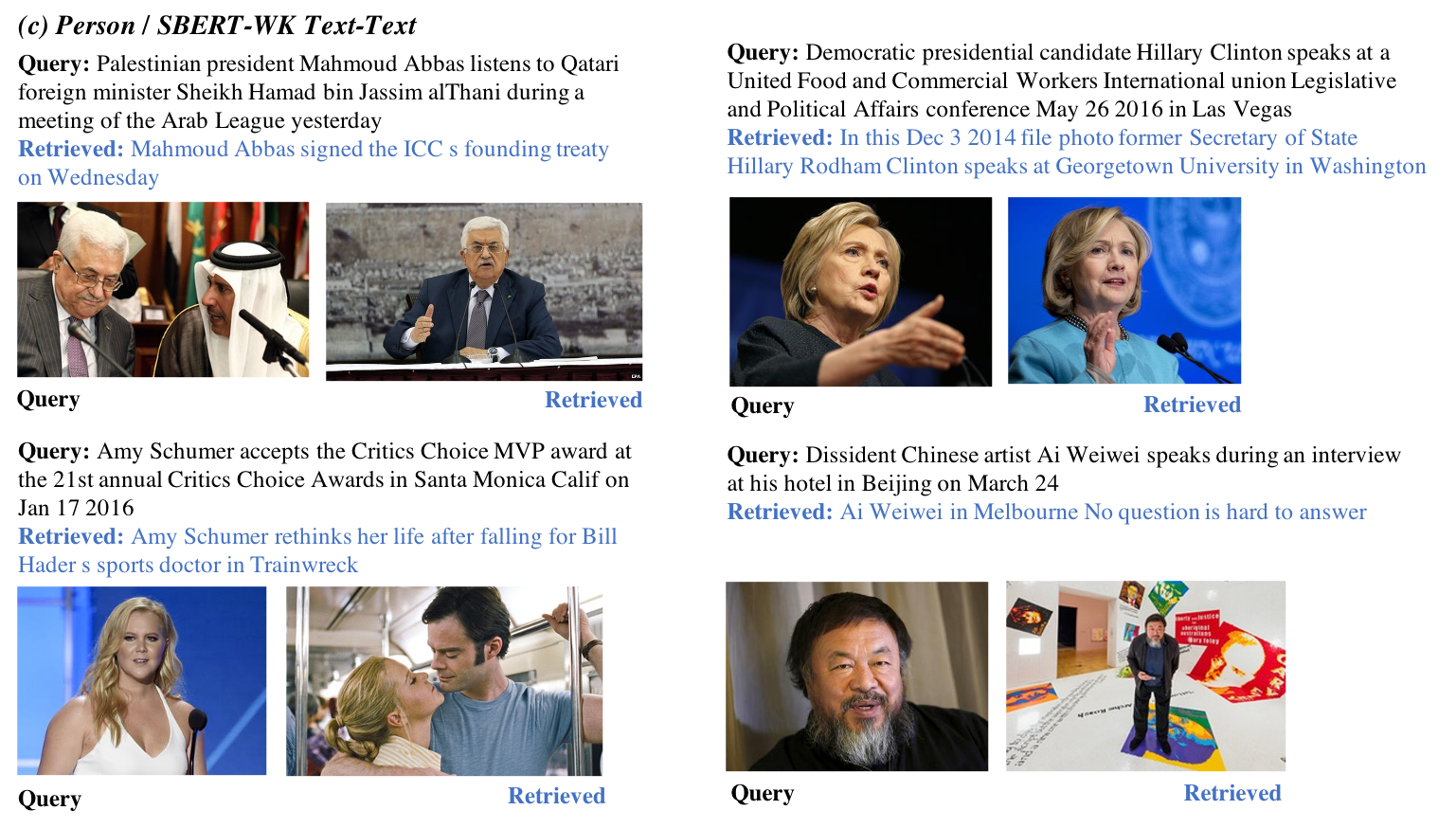}
\includegraphics[width=16cm]{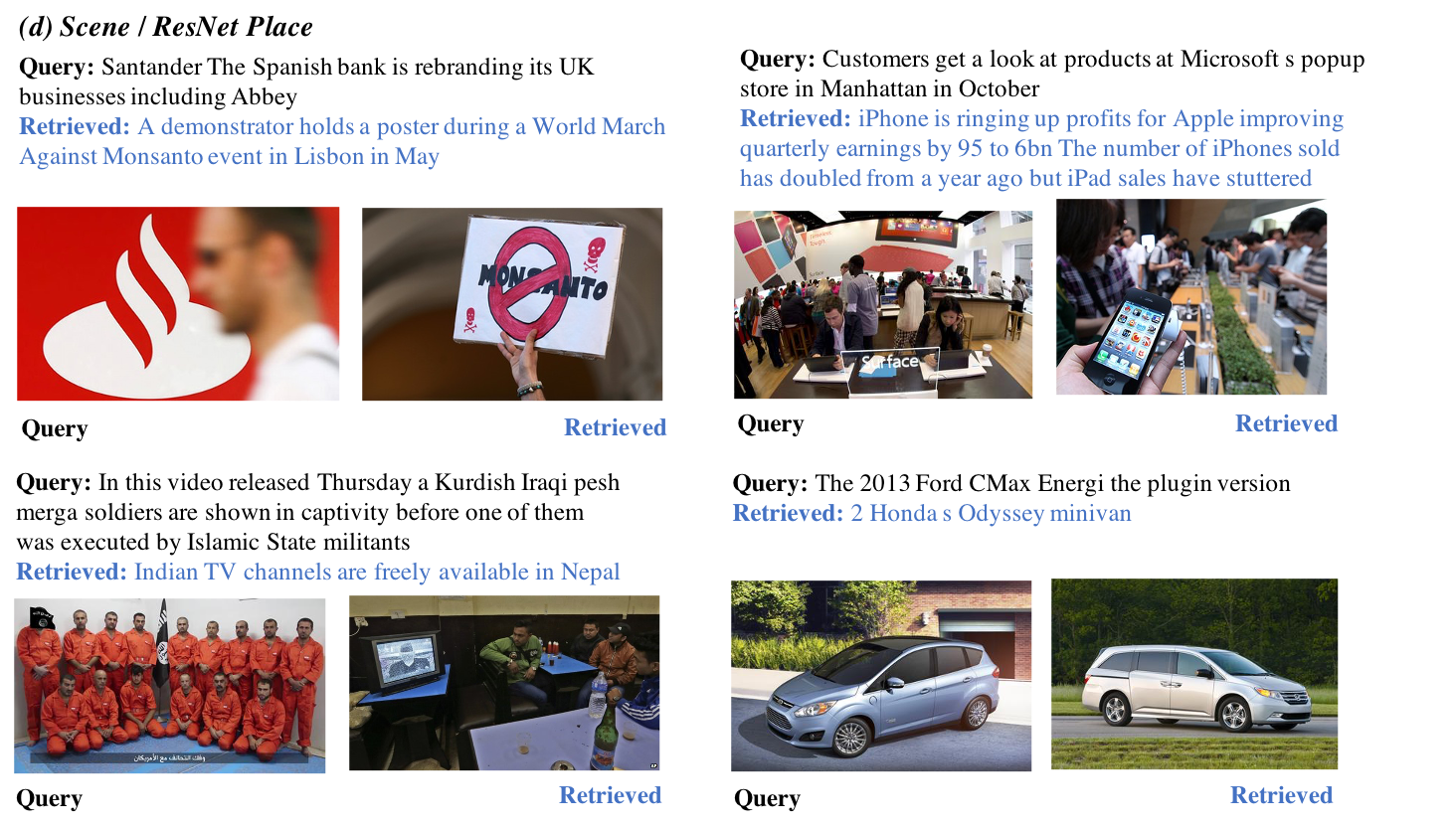}
\caption{Randomly selected (c), (d) samples from the train/val/test samples of each respective split.}
\label{fig:supplemental-random-2}
\vspace{-0.8cm}
\end{center}
\end{figure}

\twocolumn